\documentclass{article}
\usepackage{ijcai19}
\usepackage{url}
\usepackage{latexsym} 
\usepackage[english]{babel}
\usepackage[ruled, linesnumbered, noend]{algorithm2e}
\usepackage[small]{caption}
\usepackage[utf8]{inputenc}
\usepackage{amsfonts}
\usepackage{amssymb}
\usepackage{bm}
\usepackage{enumitem}
\usepackage{graphicx}
\usepackage{mathrsfs}
\usepackage{multirow}
\usepackage{soul}
\usepackage{subfigure}
\usepackage{times}
\usepackage{xcolor}
\usepackage{tabu}
\usepackage{array}
\usepackage{booktabs}
\usepackage{siunitx}
\usepackage{amsmath} 
\usepackage[hidelinks]{hyperref}
\usepackage{subfigure}
\usepackage[labelfont=bf]{caption}

\title{Story Ending Prediction by Transferable BERT}

\author{
Zhongyang Li\and
Xiao Ding\And
Ting Liu\footnote{Contact Author}\\
\affiliations
Research Center for Social Computing and Information Retrieval, Harbin Institute of Technology\\
\emails
\{zyli, xding, tliu\}@ir.hit.edu.cn
}

\begin{document}
\maketitle

\setlength{\belowcaptionskip}{-4.5pt}

\begin{abstract}
Recent advances, such as GPT and BERT, have shown success in incorporating a pre-trained transformer language model and fine-tuning operation to improve downstream NLP systems. However, this framework still has some fundamental problems in effectively incorporating supervised knowledge from other related tasks. In this study, we investigate a transferable BERT (TransBERT) training framework, which can transfer not only general language knowledge from large-scale unlabeled data but also specific kinds of knowledge from various semantically related supervised tasks, for a target task. Particularly, we propose utilizing three kinds of transfer tasks, including natural language inference, sentiment classification, and next action prediction, to further train BERT based on a pre-trained model. This enables the model to get a better initialization for the target task. We take story ending prediction as the target task to conduct experiments. The final result, an accuracy of 91.8\%, dramatically outperforms previous state-of-the-art baseline methods. Several comparative experiments give some helpful suggestions on how to select transfer tasks. Error analysis shows what are the strength and weakness of BERT-based models for story ending prediction.

\end{abstract}

\section{Introduction}
Story ending prediction, also known as the Story Cloze Test (SCT)~\cite{mostafazadeh2016corpus}, is an open task for evaluating story comprehension. This task requires a model to select the right ending from two candidate endings (one is wrong and the other is right) given a story context. The goal behind SCT is to require systems to perform deep language understanding and commonsense reasoning for successful narrative understanding, which is essential for Artificial Intelligence. There have been a variety of models trying to solve SCT so far~\cite{schwartz2017story,chaturvedi2017story,zhou2019story,li2018constructing}. However, these studies did not achieve very salient progress compared with the human performance, demonstrating the hardness of this task. Until very recently, GPT~\cite{radford2018improving} and BERT~\cite{devlin2018bert} have shown that a two-stage framework --- pre-training a language model on large-scale unsupervised corpora and fine-tuning on target tasks --- can bring promising improvements to various natural language understanding tasks, such as reading comprehension~\cite{radford2018improving} and natural language inference (NLI)~\cite{devlin2018bert}. Benefiting from these advances, the SCT performance has been pushed to a new level~\cite{radford2018improving}, though there is still a gap with the human performance.

\begin{figure} \small
    \centering
    \includegraphics[width=0.98\columnwidth]{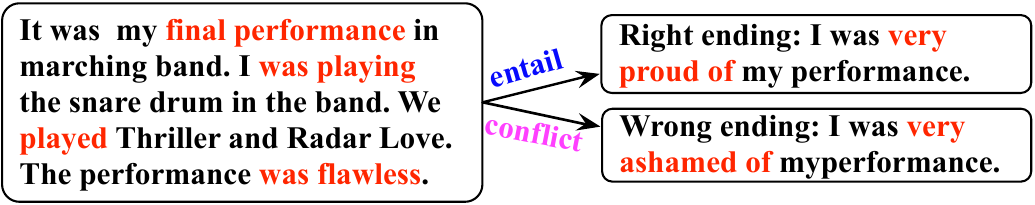}
    \vspace{-0.15cm}
    \caption{This figure shows a typical example from the development set of Story Cloze Test. There is an obvious entailment relation between the story context and the right ending, and a contradiction relation between the context and the wrong ending.}
    \label{fig:task}
\end{figure}

However, we argue that the general knowledge obtained from \textit{unsupervised} language model pre-training is not sufficient for learning a set of perfect initial parameters for every target task. Inspired by transfer learning techniques~\cite{pan2010survey}, we consider incorporating \textit{supervised} knowledge into this conventional pre-training framework to find a better initialization for the target task. Nevertheless, there still remain two fundamental problems that should be addressed:
\begin{itemize}[leftmargin=*]
\item How can the pre-training framework better utilize supervised knowledge?
\item What basic rules need to follow to find appropriate supervised knowledge for a target task?
\end{itemize}

Recently,~\cite{phang2018sentence} gave a possible solution for the first question. With a lot of crossing experiments over four intermediate tasks and nine GLUE tasks~\cite{wang2018glue}, they demonstrate that further pre-training on supervised datasets can improve the performance of GPT on downstream tasks. The MT-DNN model~\cite{liu2019multi} also tries to answer the first question by incorporating the multi-task learning framework into BERT. However, we still have no idea for answering the second challenging question from their experiments.

In this study, we take SCT as an example and try to answer the above two challenging questions through extensive experiments. We follow the idea from~\cite{phang2018sentence} and present a three-stage transferable BERT (TransBERT) framework to transfer knowledge from semantically related tasks for SCT. As shown in Figure \ref{fig:task}, the reader can easily find that the story context entails the right story ending. In contrast, the story context conflicts with the wrong ending. This suggests that the SCT task has a strong correlation with NLI. In addition, we also notice that a lot of candidate story endings in SCT are about describing human mental states and the next action following the story context. Hence, we propose utilizing three semantically related supervised tasks, including NLI, sentiment classification, and next action prediction to further pre-train the BERT model. Then the model is fine-tuned with minimal task-specific parameters to solve SCT. 

This paper makes the following three contributions:
\begin{itemize}[leftmargin=*]
\item This study presents a TransBERT framework which enables the BERT model to transfer knowledge from both unsupervised corpora and existing supervised tasks.
\item We achieve new state-of-the-art results on the widely used SCT\_v1.0 dataset and recently revised SCT\_v1.5 blind test dataset, which are much closer to the human performance.
\item Based on extensive comparative experiments, we give some helpful suggestions on how to select transfer tasks to improve BERT. Error analysis shows what are the strength and weakness of BERT-based models for SCT.
\end{itemize}

\section{Background}
Language model pre-training has shown to be very effective for learning universal language representations by leveraging large amounts of unlabeled data. Some of the most prominent models are ELMo~\cite{peters2018deep}, GPT~\cite{radford2018improving}, and BERT~\cite{devlin2018bert}. Among these, ELMo uses a bidirectional LSTM architecture, GPT exploits a left-to-right transformer architecture, while BERT uses the bidirectional transformer architecture. There are two existing strategies for applying pre-trained language models to downstream tasks: feature-based and fine-tuning. The feature-based approach, such as ELMo, uses task-specific architectures that include the pre-trained representations as input features. The fine-tuning approaches, such as GPT and BERT, introduce minimal task-specific parameters and train on the downstream tasks by jointly fine-tuning the pre-trained parameters and task-specific parameters. This two-stage framework has been demonstrated to be very effective in various natural language processing tasks, such as reading comprehension~\cite{radford2018improving} and NLI~\cite{devlin2018bert}.

\begin{figure} \small
    \centering
    \includegraphics[width=0.7\columnwidth]{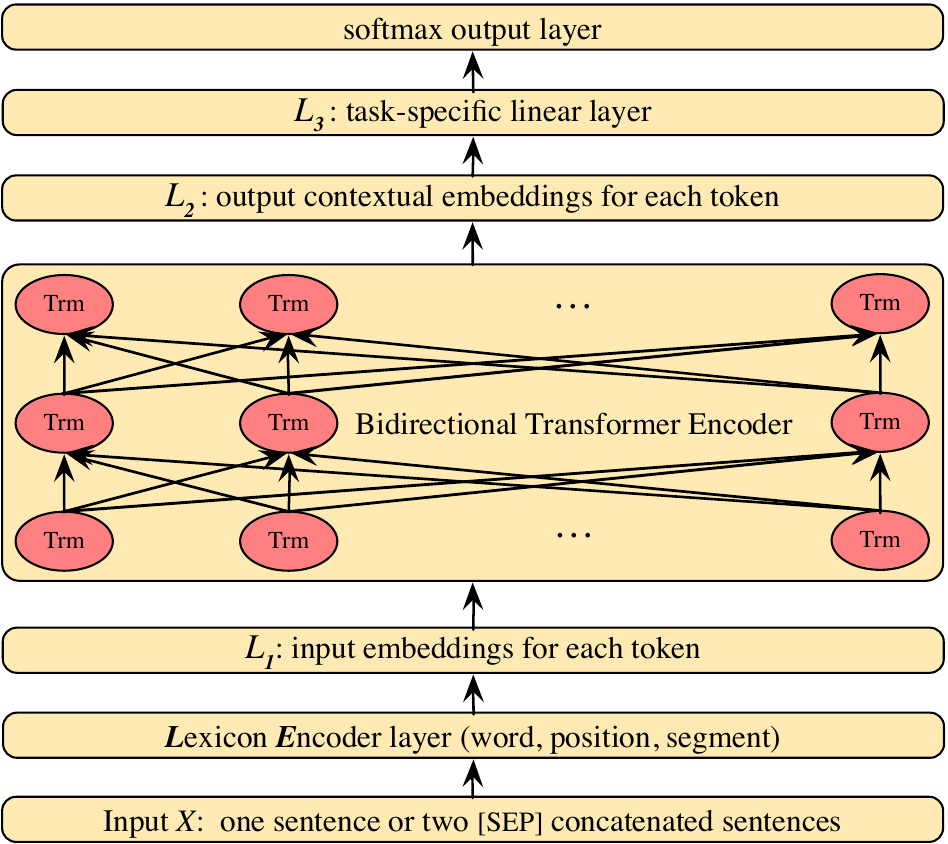}
    \vspace{-0.15cm}
    \caption{The BERT model has a lexicon encoder ($L_1$), a bidirectional transformer encoder ($L_2$), and a task specific linear layer ($L_3$).}
    \label{fig:bert}
\end{figure}

In this paper, our TransBERT training framework is based on the BERT encoder~\cite{devlin2018bert}, which exploits transformer block~\cite{vaswani2017attention} as the basic computational unit. 
Here, we describe the main components of the BERT encoder shown in Figure \ref{fig:bert}. 

The input $X$, which is a word sequence (either a sentence or two sentences concatenated together) is first represented as a sequence of input embeddings, one for each word, in $L_1$. Then the BERT encoder captures the contextual information for each word via self-attention and generates a sequence of output contextual embeddings in $L_2$. 

\textbf{Lexicon Encoder} ($L_1$): The input $X=\{x_1,...,x_n\}$ is a sequence of tokens of length $n$. The first token $x_1$ is always a special [CLS] token. If $X$ is concatenated by two sentences $X_1$ and $X_2$, they will be separated by a special token [SEP]. The lexicon encoder maps $X$ into a sequence of input embeddings, one for each token, constructed by summing the corresponding word, segment, and position embeddings. 

\textbf{Bidirectional Transformer Encoder} ($L_2$): BERT uses a multilayer bidirectional transformer encoder~\cite{vaswani2017attention} to map the input embeddings from $L_1$ into a sequence of contextual embeddings $V \in \mathbb{R}^{d\bm\cdot n}$ ($d$ is the word embedding size). The BERT model~\cite{devlin2018bert} learns the lexicon encoder and transformer encoder parameters by language model pre-training, and applies it to each downstream task by fine-tuning with minimal task-specific parameters ($L_3$).

Suppose $v_1$ is the output contextual embedding of the first token [CLS], which can be seen as the semantic representation of the whole input $X$. Take the NLI task as an example, the probability that $X$ is labeled as class $c$ (i.e., the entailment) is computed by a logistic regression with softmax:

\begin{equation*} \small
P(c|X)=\text{softmax}(W^\top_{\text{NLI}}\bm\cdot v_1)
\end{equation*}
where $W_{\text{NLI}}$ is the task-specific parameter matrix in $L_3$. 

For the task of SCT, just take the whole story context and a candidate ending as an input sentence pair, and get the output score $S$ via the BERT model. The right ending can be selected by comparing the two output scores $S_{r}$ and $S_{w}$, and choosing the ending with a higher score as the answer.

\section{The TransBERT Training Framework}
\begin{figure} \small
    \centering
    \includegraphics[width=0.95\columnwidth]{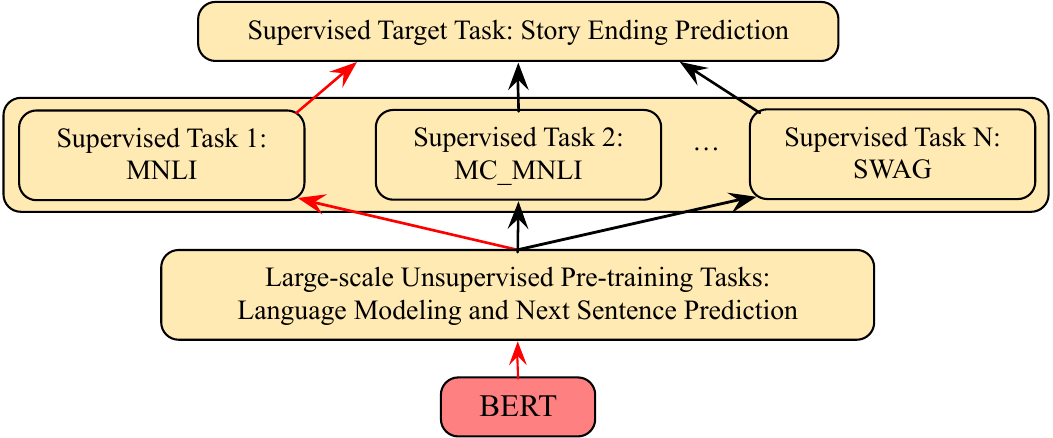}
    \vspace{-0.15cm}
    \caption{The three-stage TransBERT training framework. In this framework, we only care about the performance of the target task. The BERT model walk through a single path from bottom to the top, such as the red path shown in the figure. Hence, the model utilizes one kind of supervised knowledge each time.}
    \label{fig:model}
\end{figure}

Figure \ref{fig:model} shows the three-stage TransBERT training framework. The bottom task is unsupervised pre-training with language modeling and other related tasks, such as the next sentence prediction. In the middle of the architecture are various semantically target-related supervised tasks, which are used to further pre-train the pre-trained BERT encoder. We call such a supervised task as a \textit{Transfer Task}. On the top is the target task, specifically, SCT in this paper. The three corresponding stages can be summarized as unsupervised pre-training, supervised pre-training, and supervised fine-tuning.

\subsection{Transfer Tasks for SCT}

We believe that only when the source and the target tasks are semantically associated with each other, then the source task can be used as a transfer task. In other words, they need to share common knowledge, and this knowledge can be exploited to solve both of them.

\begin{figure} \small
    \centering
    \includegraphics[width=0.95\columnwidth]{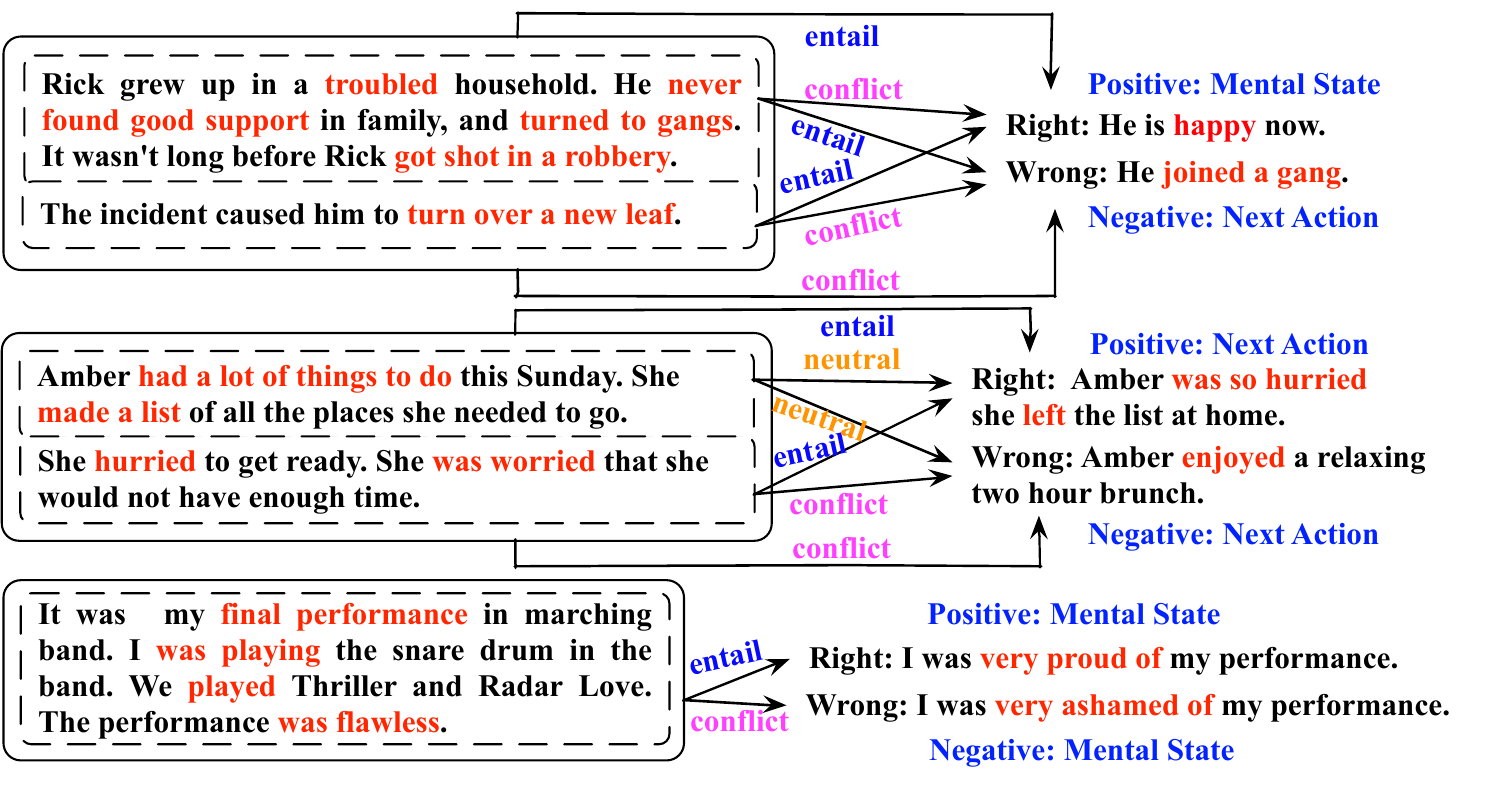}
    \vspace{-0.15cm}
    \caption{Three typical examples from the development set, which are annotated with entailment, mental states and actions information.}
    \label{fig:examples}
\end{figure}

Here, we give more intuitions why we choose NLI, sentiment classification, and next action prediction as the transfer tasks for SCT. Elementary analysis of randomly selected examples suggests that there are three typical story evolvement styles: 1. The preceding part of the context entails the wrong ending while conflicts the right ending, the succeeding part is just the opposite; 2. The preceding part of the story context has a neutral relation to both the right and wrong ending, while the succeeding part entails the right and conflicts with the wrong ending; 3. The whole story context consistently entails the right and conflicts with the wrong ending. Naturally, our intuition is that a model that can well solve the NLI task tends to have a good performance on SCT. In addition, a lot of stories especially the story endings describe human mental states or the next action following the story context. Hence, we suppose that a model that can handle the sentiment or predict the next action well, tends to have a good performance on SCT. Figure \ref{fig:examples} shows three typical examples from the development set of SCT\_v1.0, which are annotated with entailment, mental states, and actions information.

\subsubsection{Natural Language Inference}
Given a premise-hypothesis pair, the goal of NLI is to predict whether the hypothesis has an entailment, a contradiction or a neutral relation with the premise. 

\begin{itemize}[leftmargin=*]
\item \textbf{SNLI} (Stanford Natural Language Inference) dataset contains 570k human annotated sentence pairs, in which the premises are drawn from captions of Flickr30 corpus and hypotheses are manually annotated~\cite{bowman2015large}.

\item \textbf{MNLI} (Multi-Genre Natural Language Inference) is a 410k crowd-sourced multi-genre entailment classification dataset~\cite{williams2018broad}. 

\item \textbf{MC\_NLI} stands for Multiple-Choice Natural Language Inference. This dataset is a recast version of the MNLI dataset. Given a premise, we construct three kinds of hypothesis pairs: \{entailment, neutral\}, \{entailment, contradiction\}, and \{neutral, contradiction\}. The problem is to choose the entailment, entailment, and neutral hypothesis as the `right' hypothesis from the three kinds of hypothesis pairs, respectively. This dataset is used to investigate whether the transfer task having the same problem definition with the target task can provide additional benefits.

\end{itemize}

\subsubsection{Sentiment Classification}
\begin{itemize}[leftmargin=*]
\item \textbf{IMDB}~\cite{maas2011learning} contains 25K polar movie reviews for a binary sentiment classification.

\item \textbf{Twitter} is a sentiment classification dataset\footnote{http://help.sentiment140.com/} containing 1.6M tweets, which are labeled with positive and negative sentiment polarity labels.

\end{itemize}
\subsubsection{Next Action Prediction}

\textbf{SWAG} (Situations With Adversarial Generations) contains 113k sentence-pair completion examples that evaluate commonsense inference~\cite{zellers2018swag}. Given a sentence from a video captioning dataset, the task is to decide among four choices the most plausible continuation.

\subsection{Training Process}
The training procedure of TransBERT consists of three stages: unsupervised pre-training, supervised pre-training, and supervised fine-tuning. 

The first stage follows the procedure of the BERT model~\cite{devlin2018bert}. The parameters of the lexicon encoder and transformer encoder are learned using two unsupervised prediction tasks: masked language modeling and next sentence prediction. This stage allows the model to capture general knowledge and representations about the language. In this study, we use the publicly released pre-trained BERT models~\cite{devlin2018bert}. 

In the second stage, we apply the pre-trained BERT model from the first stage on various supervised tasks proposed above. For each task, minimal task-specific parameters will be introduced. These parameters will be updated jointly with the parameters of the lexicon encoder and Transformer encoder. When the model achieves the best performance on the corresponding development dataset, we save the parameters of the lexicon encoder and Transformer encoder. This stage enables the model to transfer different task-specific knowledge from various supervised tasks, and get a better model initialization for the target task. Finally, the model is fine-tuned to solve SCT with new task-specific parameters, similar to the second stage.

We train each transfer task and the SCT with 3 epochs monitoring on the development set, using a cross-entropy objective\footnote{All of our experiments are based on 
https://github.com/huggingface/pytorch-pretrained-BERT.}. Other hyper parameters follow~\cite{devlin2018bert}.

\section{Evaluation}
We evaluate the effectiveness of our model by comparing with several state-of-the-art baseline methods. Accuracy (\%) of choosing the right ending is used as the evaluation metric.

\subsection{Baselines}
We compare our model with the following baseline methods. To the best of our knowledge, most of the recent advances on Story Cloze Test are listed here.
\begin{itemize}[leftmargin=*]

\item \textbf{DSSM}~\cite{huang2013learning} measures the cosine similarity between the vector representations of the story context and the candidate endings.
\item \textbf{CGAN}~\cite{wang2017conditional} encodes the story context and each ending by GRUs and computes an entail score. 

\item \textbf{HBiLSTM}~\cite{cai2017pay} uses LSTM to encode both the word sequence and the sentence sequence, and get the modified context vector via attention.

\item \textbf{Msap}~\cite{schwartz2017story} trains a logistic regression model which uses language model and stylistic features, including length, word n-grams, and character n-grams.


\item \textbf{HCM}~\cite{chaturvedi2017story} trains a logistic regression model which combines features from event sequence, sentiment-trajectory, and topic consistency of the story.


\item \textbf{HintNet}~\cite{zhou2019story} exploits the hints which can be obtained by comparing two candidate endings to help select the correct story ending.

\item \textbf{SeqMANN}~\cite{li2018multi} uses a multi-attention neural network and introduces semantic sequence information extracted from SemLM as external knowledge. 

\item \textbf{ISCK}~\cite{chen2018incorporating} is a neural model that integrates narrative sequence, sentiment evolution, and commonsense knowledge.

\item \textbf{GPT}~\cite{radford2018improving} and \textbf{BERT}~\cite{devlin2018bert} both can solve the SCT by pre-training a language model using a multilayer transformer on open domain unlabeled corpora, followed by discriminative fine-tuning.

\end{itemize}

\subsection{Dataset}
\begin{table} \small
    \centering
    \scalebox{0.95}{
    \begin{tabular} {l c c c}
        \toprule
        \textbf{Dataset}&\textbf{Training}& \textbf{Development}& \textbf{Test}  \\
        \midrule
        SCT\_v1.0& 1,771 & 100 & 1,871 \\
        SCT\_v1.5& 1,871 & 1,571 & 1,571 \\
       \bottomrule
    \end{tabular}}
    \vspace{-0.15cm}
    \caption{Statistics of the datasets used in our experiments.}
    \label{tab:dataset}
\end{table}

To evaluate the effectiveness of our method, we experiment on two-version SCT datasets. SCT\_v1.0~\cite{mostafazadeh2016corpus} is the widely used version. It contains 98,162 five-sentence coherent stories in the training dataset (a large unlabeled stories dataset), 1,871 four-sentence story contexts along with a right ending and a wrong ending in the development and test datasets, respectively. Here we only use the development and test datasets, and split development set into 1,771 instances for training and 100 instances for development purposes. SCT\_v1.5~\cite{sharma2018tackling} is a recently released revised version in order to overcome the human-authorship biases~\cite{schwartz2017story} discovered in SCT\_v1.0. It contains 1,571 four-sentence story contexts along with a right ending and a wrong ending in the development and the blind test datasets, respectively. Here we use the 1,871 SCT\_v1.0 test dataset for training purpose.

Actually, with the released SCT\_v1.5 dataset, this paper treats the SCT\_v1.0 as a development dataset, while treats the whole SCT\_v1.5 as the real test dataset. The detailed dataset statistics are shown in Table \ref{tab:dataset}.

\section{Results and Analysis}

\begin{table} \small
    \centering
    \scalebox{0.95}{
    \begin{tabular*} {0.4\textwidth} {l c} 
        \toprule
        \textbf{Methods}& \textbf{Accuracy (\%)}  \\
        \toprule
        $\text{BERT}_{\text{BASE}}$ (multilingual, uncased) & 75.9 \\
        $\text{BERT}_{\text{BASE}}$ (multilingual, cased) & 80.2 \\
        $\text{BERT}_{\text{BASE}}$ (monolingual, cased) & 87.4 \\
        $\text{BERT}_{\text{BASE}}$ (monolingual, uncased) & 88.1 \\
        $\text{BERT}_{\text{LARGE}}$ (monolingual, uncased) & 89.2 \\
        $\text{BERT}_{\text{LARGE}}$ (monolingual, cased) & \textbf{90.0} \\
        \bottomrule
    \end{tabular*}}
    \vspace{-0.15cm}
    \caption{Experimental results with all the publicly released pre-trained BERT models on SCT\_v1.0 test dataset. `Uncased' means all words in the training corpus will be transformed into lower case form. `Cased' means keeping all the words in their original form.}
    \label{tab:bert_versions}
\end{table}

There are several pre-trained BERT models available to the community~\cite{devlin2018bert}. They differ in how many layers and parameters are used in the model (the basic version has 12-layer transformer blocks, 768 hidden-size, and 12 self-attention heads, totally 110M parameters; the large version has 24-layer transformer blocks, 1024 hidden-size, and 16 self-attention heads, totally 340M parameters), and what kind of datasets are used to pre-train the model (multilingual or monolingual). We first conduct several comparative experiments on SCT\_v1.0 dataset to study the effectiveness of different BERT versions. Results are shown in Table \ref{tab:bert_versions}. We find that the two multilingual models perform much worse than the monolingual models. An uncased $\text{BERT}_{\text{BASE}}$ performs better than the cased $\text{BERT}_{\text{BASE}}$, but a cased $\text{BERT}_{\text{LARGE}}$ is better than the uncased $\text{BERT}_{\text{LARGE}}$. The reasons are that the multilingual BERT model doesn't improve the performance on the monolingual SCT english dataset; the $\text{BERT}_{\text{LARGE}}$ model can handle a larger cased vocabulary with much more parameters but the $\text{BERT}_{\text{BASE}}$ model cannot.

In the following experiments, $\text{BERT}_{\text{BASE}}$ refers to the uncased monolingual version of $\text{BERT}_{\text{BASE}}$ model, and $\text{BERT}_{\text{LARGE}}$ refers to the cased monolingual version of $\text{BERT}_{\text{LARGE}}$ model.

\subsection{Overall Results}
\begin{table} \small
    \centering
    \scalebox{0.99}{
    \begin{tabular*} {0.4\textwidth} {l c} 
        \toprule
        \textbf{Method}& \textbf{Accuracy (\%)}  \\
        \toprule
        DSSM~\cite{huang2013learning} & 58.5 \\
        CGAN~\cite{wang2017conditional}& 60.9 \\
        HBiLSTM~\cite{cai2017pay}& 74.7 \\
        Msap~\cite{schwartz2017story}& 75.2 \\
        HCM~\cite{chaturvedi2017story}& 77.6 \\
        HintNet~\cite{zhou2019story}& 79.2 \\
        SeqMANN~\cite{li2018multi}& 84.7 \\
        GPT~\cite{radford2018improving}& 86.5 \\
        ISCK~\cite{chen2018incorporating}& 87.6 \\
        $\text{BERT}_{\text{BASE}}$ (Our Implementation)& 88.1 \\
        $\text{BERT}_{\text{LARGE}}$ (Our Implementation)& \textbf{90.0} \\
        \hline
        $\text{BERT}_{\text{BASE}}$ + SNLI (Ours) & 85.9 \\
        $\text{BERT}_{\text{BASE}}$ + IMDB (Ours) & 87.6 \\
        $\text{BERT}_{\text{BASE}}$ + SWAG (Ours) & 88.6 \\
        $\text{BERT}_{\text{BASE}}$ + Twitter (Ours) & 88.7 \\
        $\text{BERT}_{\text{BASE}}$ + MC\_MNLI (Ours) & 89.5 \\
        $\text{BERT}_{\text{BASE}}$ + MNLI (Ours) & 90.6 \\
        $\text{BERT}_{\text{LARGE}}$ + MNLI (Ours) & \textbf{91.8} \\
        \hline
        Human~\cite{mostafazadeh2016corpus}& 100.0 \\
        \bottomrule
    \end{tabular*}}
    \vspace{-0.15cm}
    \caption{Experimental results of story ending prediction on SCT\_v1.0 test dataset. Differences between our best method and all baseline methods are significant ($p < 0.01$) using t-test. $\text{BERT}_{\text{LARGE}}$ + MNLI also gets the SOTA performance of \textbf{90.3\%} on the newly released SCT\_v1.5 blind test dataset, which is not shown in this table.}
    \label{tab:overall_results10}
\end{table}

Table \ref{tab:overall_results10} shows the overall experimental results on SCT\_v1.0 test dataset. The best previously reported result is from ISCK~\cite{chen2018incorporating}, which is an accuracy of 87.6\%. We implemented the same BERT model as~\cite{devlin2018bert} and got the best baseline results on SCT, which are 88.1\% and 90.0\% from $\text{BERT}_{\text{BASE}}$ and $\text{BERT}_{\text{LARGE}}$ models, respectively. This is because the BERT model can obtain general language knowledge from pre-training. From Table \ref{tab:overall_results10} we can also find that most of our transfer tasks can further improve BERT, except SNLI and IMDB. The MNLI-enhanced BERT models achieved the best accuracies of 90.5\% and 91.8\%, which are the new state-of-the-art performances on SCT\_v1.0. This is because our method can learn task-specific knowledge from transfer tasks, which is helpful for SCT, and MNLI is the most informative task.

Table \ref{tab:overall_results10} also shows some interesting results. Comparing the SNLI and MNLI-enhanced BERT models, we find that though NLI can help SCT intuitively, the data source still plays an important role. MNLI is a multi-genre dataset, while SNLI data is from the specific image caption domain. Hence, the MNLI tends to help the open domain SCT but SNLI does not. Comparing the IMDB and Twitter-enhanced BERT models, we can get similar conclusions that the open domain Twitter can improve the performance of BERT on SCT, while the specific domain IMDB hurts the model's performance. Comparing the MC\_MNLI and MNLI-enhanced BERT models, we find that MNLI helps more for SCT (multiple choice task). Hence, we can get the conclusion that the transfer task doesn't need to have the same problem definition as the target task. This is mainly because the model can get a better knowledge about entailment when NLI is formulated as a classification task (MNLI), other than a multiple choice task (MC\_MNLI).

\subsection{Comparative Experiments}
Several comparative experiments are conducted to investigate some fine-grained aspects.

\subsubsection{Whether All Four Sentences in the Story Context Are Useful for BERT to Choose the Right Ending?}

Different from NLI and SWAG, in which there are only two sentences in an instance, the SCT has a longer four-sentence context. We experiment to investigate whether the BERT-based models can make full use of the long story context. Experimental results are shown in Table \ref{tab:story_context}. We find that all the sentences in the story context are useful and the BERT-based models can make full use of them to infer the correct ending. This is mainly because the BERT-based models have the ability to handle long distance dependency with the self-attention mechanism \cite{vaswani2017attention}.

\begin{table} \small
    \centering
    \scalebox{0.97}{
    \begin{tabular*} {0.39\textwidth} {l c} 
        \toprule
        \textbf{Method}& \textbf{Accuracy (\%)}  \\
        \toprule
        $\text{BERT}_{\text{BASE}}$ (ending only) & 77.9 \\
        $\text{BERT}_{\text{BASE}}$ (4) & 86.4 \\
        $\text{BERT}_{\text{BASE}}$ (3,4) & 87.4 \\
        $\text{BERT}_{\text{BASE}}$ (2,3,4) & 87.7 \\
        $\text{BERT}_{\text{BASE}}$ (1,2,3,4) & \textbf{88.1} \\
        \hline
        $\text{BERT}_{\text{BASE}}$ + MNLI (ending only) & 78.3 \\
        $\text{BERT}_{\text{BASE}}$ + MNLI (4) & 88.5 \\
        $\text{BERT}_{\text{BASE}}$ + MNLI (3,4) & 88.7 \\
        $\text{BERT}_{\text{BASE}}$ + MNLI (2,3,4) & 88.6 \\
        $\text{BERT}_{\text{BASE}}$ + MNLI (1,2,3,4) & \textbf{90.6} \\
        \bottomrule
    \end{tabular*}}
    \vspace{-0.15cm}
    \caption{Experimental results with different sentences combination as the story context. (3,4) means only the third and the fourth sentences are used as the story context, and other settings are similar.}
    \label{tab:story_context}
\end{table}

\subsubsection{Explore the Effectiveness of Different MNLI Categories}
\begin{table} \small
    \centering
    \scalebox{0.97}{
    \begin{tabular*} {0.37\textwidth} {l c} 
        \toprule
        \textbf{Method}& \textbf{Accuracy (\%)}  \\
        \toprule
        $\text{BERT}_{\text{BASE}}$ & 88.1 \\
        $\text{BERT}_{\text{BASE}}$ + MNLI (EN only) & 86.2 \\
        $\text{BERT}_{\text{BASE}}$ + MNLI (NC only) & 88.8 \\
        $\text{BERT}_{\text{BASE}}$ + MNLI (EC only) & 89.2 \\
        $\text{BERT}_{\text{BASE}}$ + MNLI & \textbf{90.6} \\
        \bottomrule
    \end{tabular*}}
    \vspace{-0.15cm}
    \caption{Experimental results with different natural language inference categories on SCT\_v1.0 test dataset. (EN only) means this setting only considers the \textbf{E}ntailment and \textbf{N}eutral realtions, with the \textbf{C}ontradiction relation filtered out.} 
    \label{tab:mnli_categories}
\end{table}

Our experiments suggest that we can achieve the best performance when using MNLI as the transfer task. But we also want to know which category among the \textbf{E}ntailment, \textbf{N}eutral and \textbf{C}ontradiction is the most informative for SCT. The results are shown in Table \ref{tab:mnli_categories}. We find that the contradiction relation is the most informative one, then entailment, and neutral the least. It's interesting that the performance even drops a lot without the contradiction. The reason is that the ability to recognize conflict endings enables the model to pick up the right ending more easily. Finally, the best performance is achieved by using all three relations together, demonstrating that each relation can help SCT from different aspects.

\subsection{Discussion and Analysis}

The MNLI-enhanced BERT models push the performance to 91.8\% and 90.3\% accuracies on SCT\_v1.0 and  SCT\_v1.5 test datasets, respectively, which are much closer to the human performance (100\%). Though very effective in natural language understanding, there are still about 9\% of the test instances that the BERT-based models cannot handle. 

First, we are curious about why MNLI can improve SCT with such a large margin. Hence, we trained a model on the MNLI dataset and directly applied it to solve the SCT task. Surprisingly, this simple method got a relatively high accuracy of 63.4\% on SCT\_v1.0 test set, even better than the DSSM and CGAN models which were trained on the SCT dataset. This demonstrates the high correlation between MNLI and SCT. We argue that the SCT task can be seen as a more complicated NLI task, where the premise is a more complex four-sentence evolving context. The goal is to find the right ending that can be entailed with a higher probability than the wrong ending, with respect to the story context.

Error analysis of the unsolved instances shows that BERT-based models make a lot of mistakes when one of the two candidate endings is about mental state while the other one describes the next action. This is mainly because BERT-based models are good at distinguishing from two homogeneous endings (e.g. both describe human mental states). But they cannot handle two heterogeneous endings well. Better models will be needed to handle this properly.

Here we try to answer the above two challenging questions:
\begin{itemize}[leftmargin=*]

\item How can the pre-training framework better utilize supervised knowledge: One way is to add a second pre-training stage to integrate knowledge from existing supervised tasks, like what the STILTs~\cite{phang2018sentence} and TransBERT do. But this method can only exploit one single supervised task each time. Another way is to pre-train the transfer tasks in a multi-task learning manner~\cite{liu2019multi} (e.g. train MNLI, Twitter, and SWAG simultaneously). But it's unknown whether this multi-task learning manner can bring more improvement to SCT, even if each of the three tasks is helpful. We leave this as future work.

\item What basic rules need to follow to find appropriate supervised knowledge for a target task: First, the transfer task and the target task need to be semantically associated with each other and share common knowledge between them. This knowledge can be exploited to solve both of them. Second, this paper explores transferring knowledge from different supervised tasks to SCT, showing that a specific domain dataset (SNLI) is not sufficient for improving an open domain target task (SCT), even though they are semantically associated with each other. Third, the transfer task doesn't need to have the same problem definition as the target task. A classification transfer task (MNLI) can help a multiple choice target task (SCT). 

\end{itemize}

\section{Related Work}

\noindent \textbf{The Story Cloze Test}
Story Cloze Test~\cite{mostafazadeh2016corpus} is a task for evaluating story understanding. Previous methods on this task can be roughly categorized into two lines: feature-based methods~\cite{schwartz2017story,chaturvedi2017story} and neural models~\cite{cai2017pay}. 

Feature-based methods for SCT~\cite{schwartz2017story} adopted shallow semantic features, such as n-grams and POS tags, and trained a linear regression model to determine whether a candidate ending is plausible. 
HCM~\cite{chaturvedi2017story} further integrated event, sentiment and topic into feature-based methods. 

Neural models~\cite{cai2017pay,zhou2019story} for SCT learn embeddings for the story context and candidate endings, and select the right ending by computing the embeddings' similarity. 
SeqMANN~\cite{li2018multi} integrated external knowledge into a multi-attention network. 
GPT~\cite{radford2018improving} pre-trained a transformer language model and fine-tuned the model to solve SCT. 
ISCK~\cite{chen2018incorporating} used a neural model that integrated narrative sequence, sentiment evolution, and commonsense knowledge.
Instead of choosing the right ending, several previous studies aimed to directly generate a reasonable ending~\cite{li2018generating}. 

Different from the previous commonsense models, we try to incorporate knowledge from other supervised tasks into the most advanced BERT representation model.

\noindent \textbf{Learning Universal Language Representations}
Language model pre-training has shown to be very effective for learning universal language representations. Among these models, ELMo~\cite{peters2018deep} and ULMFiT~\cite{howard2018universal} used a BiLSTM architecture, while GPT~\cite{radford2018improving} and BERT~\cite{devlin2018bert} utilized the transformer architecture~\cite{vaswani2017attention}. Unlike most earlier approaches, such as ELMo, where the weights of the encoder were frozen after pre-training, ULMFiT, GPT and BERT jointly fine-tuned the encoder and task-specific parameters on the downstream tasks. 

STILTs~\cite{phang2018sentence} fine-tuned a GPT model on some intermediate tasks to get better performance on the GLUE~\cite{wang2018glue} benchmark. However, they gave little analysis of this transfer mechanism. Take SCT as an example, we give some helpful suggestions and our insights on how to select transfer tasks.

\noindent \textbf{Transfer Learning and Multi-task Learning}
Transfer learning~\cite{pan2010survey} is widely adopted in the NLP community, such as dialogue system~\cite{mo2018personalizing} and text style transfer~\cite{fu2018style}.
This work is also related to multi-task learning~\cite{liurepresentation}, where multiple tasks were jointly trained to get an overall performance improvement. MT-DNN~\cite{liu2019multi} extended multi-task learning by incorporating a pre-trained BERT model, which is very close to the work of this paper.

\section{Conclusion}
In this paper, we present a three-stage training framework TransBERT, which can transfer not only general language knowledge from large-scale unlabeled data but also specific kinds of knowledge from various semantically associated supervised tasks for a target task, such as SCT. This training framework can enable a better and task-specific initialization for different target tasks, which is superior to the widely used two-stage pre-training and fine-tuning framework. The MNLI-enhanced BERT model pushes the SCT\_v1.0 task to 91.8\% accuracy, which is much closer to human performance. It also gets the SOTA performance of 90.3\% on SCT\_v1.5.

\section*{Acknowledgments}
This work is supported by the National Natural Science Foundation of China via grants 61632011, 61702137 and 61772153. 
Thanks to the reviewers' insightful comments. 

\newpage
\bibliographystyle{named}
\bibliography{ijcai19}

\begin{thebibliography}{}

\bibitem[\protect\citeauthoryear{Bowman \bgroup \em et al.\egroup
  }{2015}]{bowman2015large}
Samuel~R Bowman, Gabor Angeli, Christopher Potts, and Christopher~D Manning.
\newblock A large annotated corpus for learning natural language inference.
\newblock In {\em EMNLP}, pages 632--642, 2015.

\bibitem[\protect\citeauthoryear{Cai \bgroup \em et al.\egroup
  }{2017}]{cai2017pay}
Zheng Cai, Lifu Tu, and Kevin Gimpel.
\newblock Pay attention to the ending: Strong neural baselines for the roc
  story cloze task.
\newblock In {\em ACL}, pages 616--622, 2017.

\bibitem[\protect\citeauthoryear{Chaturvedi \bgroup \em et al.\egroup
  }{2017}]{chaturvedi2017story}
Snigdha Chaturvedi, Haoruo Peng, and Dan Roth.
\newblock Story comprehension for predicting what happens next.
\newblock In {\em EMNLP}, pages 1603--1614, 2017.

\bibitem[\protect\citeauthoryear{Chen \bgroup \em et al.\egroup
  }{2019}]{chen2018incorporating}
Jiaao Chen, Jianshu Chen, and Zhou Yu.
\newblock Incorporating structured commonsense knowledge in story completion.
\newblock {\em AAAI}, 2019.

\bibitem[\protect\citeauthoryear{Devlin \bgroup \em et al.\egroup
  }{2018}]{devlin2018bert}
Jacob Devlin, Ming-Wei Chang, Kenton Lee, and Kristina Toutanova.
\newblock Bert: Pre-training of deep bidirectional transformers for language
  understanding.
\newblock {\em arXiv preprint arXiv:1810.04805}, 2018.

\bibitem[\protect\citeauthoryear{Fu \bgroup \em et al.\egroup
  }{2018}]{fu2018style}
Zhenxin Fu, Xiaoye Tan, Nanyun Peng, Dongyan Zhao, and Rui Yan.
\newblock Style transfer in text: Exploration and evaluation.
\newblock In {\em AAAI}, 2018.

\bibitem[\protect\citeauthoryear{Howard and Ruder}{2018}]{howard2018universal}
Jeremy Howard and Sebastian Ruder.
\newblock Universal language model fine-tuning for text classification.
\newblock In {\em ACL}, volume~1, pages 328--339, 2018.

\bibitem[\protect\citeauthoryear{Huang \bgroup \em et al.\egroup
  }{2013}]{huang2013learning}
Po-Sen Huang, Xiaodong He, Jianfeng Gao, Li~Deng, Alex Acero, and Larry Heck.
\newblock Learning deep structured semantic models for web search using
  clickthrough data.
\newblock In {\em CIKM}, pages 2333--2338. ACM, 2013.

\bibitem[\protect\citeauthoryear{Li \bgroup \em et al.\egroup
  }{2018a}]{li2018multi}
Qian Li, Ziwei Li, Jin-Mao Wei, Yanhui Gu, Adam Jatowt, and Zhenglu Yang.
\newblock A multi-attention based neural network with external knowledge for
  story ending predicting task.
\newblock In {\em Coling}, pages 1754--1762, 2018.

\bibitem[\protect\citeauthoryear{Li \bgroup \em et al.\egroup
  }{2018b}]{li2018constructing}
Zhongyang Li, Xiao Ding, and Ting Liu.
\newblock Constructing narrative event evolutionary graph for script event
  prediction.
\newblock In {\em IJCAI}, pages 4201--4207, 2018.

\bibitem[\protect\citeauthoryear{Li \bgroup \em et al.\egroup
  }{2018c}]{li2018generating}
Zhongyang Li, Xiao Ding, and Ting Liu.
\newblock Generating reasonable and diversified story ending using sequence to
  sequence model with adversarial training.
\newblock In {\em Coling}, pages 1033--1043. ACL, August 2018.

\bibitem[\protect\citeauthoryear{Liu \bgroup \em et al.\egroup
  }{2015}]{liurepresentation}
Xiaodong Liu, Jianfeng Gao, Xiaodong He, Li~Deng, Kevin Duh, and Ye-yi Wang.
\newblock Representation learning using multi-task deep neural networks for
  semantic classification and information retrieval.
\newblock pages 912--921, 2015.

\bibitem[\protect\citeauthoryear{Liu \bgroup \em et al.\egroup
  }{2019}]{liu2019multi}
Xiaodong Liu, Pengcheng He, Weizhu Chen, and Jianfeng Gao.
\newblock Multi-task deep neural networks for natural language understanding.
\newblock {\em arXiv preprint arXiv:1901.11504}, 2019.

\bibitem[\protect\citeauthoryear{Maas \bgroup \em et al.\egroup
  }{2011}]{maas2011learning}
Andrew~L Maas, Raymond~E Daly, Peter~T Pham, Dan Huang, Andrew~Y Ng, and
  Christopher Potts.
\newblock Learning word vectors for sentiment analysis.
\newblock In {\em ACL}, pages 142--150. ACL, 2011.

\bibitem[\protect\citeauthoryear{Mo \bgroup \em et al.\egroup
  }{2018}]{mo2018personalizing}
Kaixiang Mo, Yu~Zhang, Shuangyin Li, Jiajun Li, and Qiang Yang.
\newblock Personalizing a dialogue system with transfer reinforcement learning.
\newblock In {\em AAAI}, 2018.

\bibitem[\protect\citeauthoryear{Mostafazadeh \bgroup \em et al.\egroup
  }{2016}]{mostafazadeh2016corpus}
Nasrin Mostafazadeh, Nathanael Chambers, Xiaodong He, Devi Parikh, Dhruv Batra,
  Lucy Vanderwende, Pushmeet Kohli, and James Allen.
\newblock A corpus and cloze evaluation for deeper understanding of commonsense
  stories.
\newblock {\em NAACL}, pages 740--750, 2016.

\bibitem[\protect\citeauthoryear{Pan and Yang}{2009}]{pan2010survey}
Sinno~Jialin Pan and Qiang Yang.
\newblock A survey on transfer learning.
\newblock {\em TKDE}, 22(10):1345--1359, 2009.

\bibitem[\protect\citeauthoryear{Peters \bgroup \em et al.\egroup
  }{2018}]{peters2018deep}
Matthew Peters, Mark Neumann, Mohit Iyyer, Matt Gardner, Christopher Clark,
  Kenton Lee, and Luke Zettlemoyer.
\newblock Deep contextualized word representations.
\newblock In {\em NAACL}, volume~1, pages 2227--2237, 2018.

\bibitem[\protect\citeauthoryear{Phang \bgroup \em et al.\egroup
  }{2018}]{phang2018sentence}
Jason Phang, Thibault F{\'e}vry, and Samuel~R Bowman.
\newblock Sentence encoders on stilts: Supplementary training on intermediate
  labeled-data tasks.
\newblock {\em arXiv preprint arXiv:1811.01088}, 2018.

\bibitem[\protect\citeauthoryear{Radford \bgroup \em et al.\egroup
  }{2018}]{radford2018improving}
Alec Radford, Karthik Narasimhan, Tim Salimans, and Ilya Sutskever.
\newblock Improving language understanding by generative pre-training.
\newblock 2018.

\bibitem[\protect\citeauthoryear{Schwartz \bgroup \em et al.\egroup
  }{2017}]{schwartz2017story}
Roy Schwartz, Maarten Sap, Ioannis Konstas, Leila Zilles, Yejin Choi, and
  Noah~A Smith.
\newblock Story cloze task: Uw nlp system.
\newblock In {\em LSDSem}, pages 52--55, 2017.

\bibitem[\protect\citeauthoryear{Sharma \bgroup \em et al.\egroup
  }{2018}]{sharma2018tackling}
Rishi Sharma, James Allen, Omid Bakhshandeh, and Nasrin Mostafazadeh.
\newblock Tackling the story ending biases in the story cloze test.
\newblock In {\em ACL}, volume~2, pages 752--757, 2018.

\bibitem[\protect\citeauthoryear{Vaswani \bgroup \em et al.\egroup
  }{2017}]{vaswani2017attention}
Ashish Vaswani, Noam Shazeer, Niki Parmar, Jakob Uszkoreit, Llion Jones,
  Aidan~N Gomez, {\L}ukasz Kaiser, and Illia Polosukhin.
\newblock Attention is all you need.
\newblock In {\em NIPS}, pages 5998--6008, 2017.

\bibitem[\protect\citeauthoryear{Wang \bgroup \em et al.\egroup
  }{2017}]{wang2017conditional}
Bingning Wang, Kang Liu, and Jun Zhao.
\newblock Conditional generative adversarial networks for commonsense machine
  comprehension.
\newblock In {\em IJCAI}, pages 4123--4129, 2017.

\bibitem[\protect\citeauthoryear{Wang \bgroup \em et al.\egroup
  }{2018}]{wang2018glue}
Alex Wang, Amanpreet Singh, Julian Michael, Felix Hill, Omer Levy, and Samuel
  Bowman.
\newblock Glue: A multi-task benchmark and analysis platform for natural
  language understanding.
\newblock In {\em EMNLP Workshop}, pages 353--355, 2018.

\bibitem[\protect\citeauthoryear{Williams \bgroup \em et al.\egroup
  }{2018}]{williams2018broad}
Adina Williams, Nikita Nangia, and Samuel Bowman.
\newblock A broad-coverage challenge corpus for sentence understanding through
  inference.
\newblock In {\em NAACL}, volume~1, pages 1112--1122, 2018.

\bibitem[\protect\citeauthoryear{Zellers \bgroup \em et al.\egroup
  }{2018}]{zellers2018swag}
Rowan Zellers, Yonatan Bisk, Roy Schwartz, and Yejin Choi.
\newblock Swag: A large-scale adversarial dataset for grounded commonsense
  inference.
\newblock In {\em EMNLP}, pages 93--104, 2018.

\bibitem[\protect\citeauthoryear{Zhou \bgroup \em et al.\egroup
  }{2019}]{zhou2019story}
Mantong Zhou, Minlie Huang, and Xiaoyan Zhu.
\newblock Story ending selection by finding hints from pairwise candidate
  endings.
\newblock {\em TASLP}, 2019.

\end{thebibliography}
\end{document}